\documentclass[letterpaper, 10 pt, conference]{ieeeconf}  % Comment this line out if you need a4paper

\IEEEoverridecommandlockouts                              % This command is only needed if 
\usepackage{graphics} % for pdf, bitmapped graphics files
\usepackage{graphicx} % for epx
\usepackage{epsfig} % for postscript graphics files
\usepackage{amsmath, bm} % assumes amsmath package installed
\usepackage{amssymb}  % assumes amsmath package installed
\usepackage{upgreek}
\usepackage{multicol}
\usepackage{multirow}
\usepackage{lipsum}  
\usepackage{breqn}
\usepackage{comment} % Block comment
\usepackage{booktabs}

\PassOptionsToPackage{hyphens}{url}\usepackage{hyperref}
\usepackage[caption=false]{subfig} % Subfigures

\usepackage{xcolor}

\usepackage[style=ieee]{biblatex}

\DeclareSourcemap{
  \maps{
    \map{
      \pertype{article}
      \step[fieldset=language, null]
      \step[fieldset=url, null]
      \step[fieldset=doi, null]
      \step[fieldset=issn, null]
      \step[fieldset=isbn, null]
      \step[fieldset=note, null]
      \step[fieldset=editor, null]
      \step[fieldset=urldate, null]
      \step[fieldset=file, null]
    }
  }
}
\DeclareSourcemap{
  \maps{
    \map{
      \pertype{inproceedings}
      \step[fieldset=language, null]
      \step[fieldset=url, null]
      \step[fieldset=doi, null]
      \step[fieldset=issn, null]
      \step[fieldset=isbn, null]
      \step[fieldset=note, null]
      \step[fieldset=editor, null]
      \step[fieldset=urldate, null]
      \step[fieldset=file, null]
    }
  }
}
\DeclareSourcemap{
  \maps{
    \map{
      \pertype{incollection}
      \step[fieldset=language, null]
      \step[fieldset=url, null]
      \step[fieldset=doi, null]
      \step[fieldset=issn, null]
      \step[fieldset=isbn, null]
      \step[fieldset=note, null]
      \step[fieldset=editor, null]
      \step[fieldset=urldate, null]
      \step[fieldset=file, null]
    }
  }
}

\title{\LARGE \bf
Unsteady aerodynamic modeling of \textit{Aerobat} using lifting line theory and Wagner's function
}

\author{Eric Sihite$^{1}$, Paul Ghanem$^{2}$, Adarsh Salagame$^{2}$, and Alireza Ramezani$^{2}$%
\thanks{$^{1}$ The author is with the Department of Aerospace, California Institute of Technology, Pasadena, CA-91125, USA. (e-mail: esihite@caltech.edu).}%
\thanks{$^{2}$ The authors are with the SiliconSynapse Laboratory, Department of Electrical and Computer Engineering, Northeastern University, Boston, MA-02119, USA. (e-mail: ghanem.p, salagame.a, a.ramezani@northeastern.edu).}%
}

\begin{document}

\maketitle
\thispagestyle{empty}
\pagestyle{empty}

%%%%%%%%%%%%%%%%%%%%%%%%%%%%%%%%%%%%%%%%%%%%%%%%%%%%%%%%%%%%%%%%%%%%%%%%%%%%%%%%
\begin{abstract}

Flying animals possess highly complex physical characteristics and are capable of performing agile maneuvers using their wings. The flapping wings generate complex wake structures that influence the aerodynamic forces, which can be difficult to model. While it is possible to model these forces using fluid-structure interaction, it is very computationally expensive and difficult to formulate. In this paper, we follow a simpler approach by deriving the aerodynamic forces using a relatively small number of states and presenting them in a simple state-space form. The formulation utilizes Prandtl's lifting line theory and Wagner's function to determine the unsteady aerodynamic forces acting on the wing in a simulation, which then are compared to experimental data of the bat-inspired robot called the \textit{Aerobat}. The simulated trailing-edge vortex shedding can be evaluated from this model, which then can be analyzed for a wake-based gait design approach to improve the aerodynamic performance of the robot.

\end{abstract}

%%%%%%%%%%%%%%%%%%%%%%%%%%%%%%%%%%%%%%%%%%%%%%%%%%%%%%%%%%%%%%%%%%%%%%%%%%%%%%%%

\section{Introduction}
\label{sec:introduction}

Flying animals possess highly complex physical characteristics and create wake structures downstream of their flight path as they flap their wings through the fluidic environment \cite{hedenstrom_bat_2015}. Bats, in particular, possess highly dexterous and flexible wings that are capable of performing agile maneuvers as they dynamically morph their wings during flight. Their arm-wing possesses over 40 modes to characterize their flapping gait and highly flexible wings \cite{riskin_quantifying_2008}. These complexities are a major source of motivation for us to take inspiration and learn from, and we look to develop unmanned aerial vehicles (UAVs) that can mimic bat's flight by using the wings as the primary source of thrust and lift generation.

\begin{figure}[t]
    \centering
    \vspace{0.1in}
    \includegraphics[width = \linewidth]{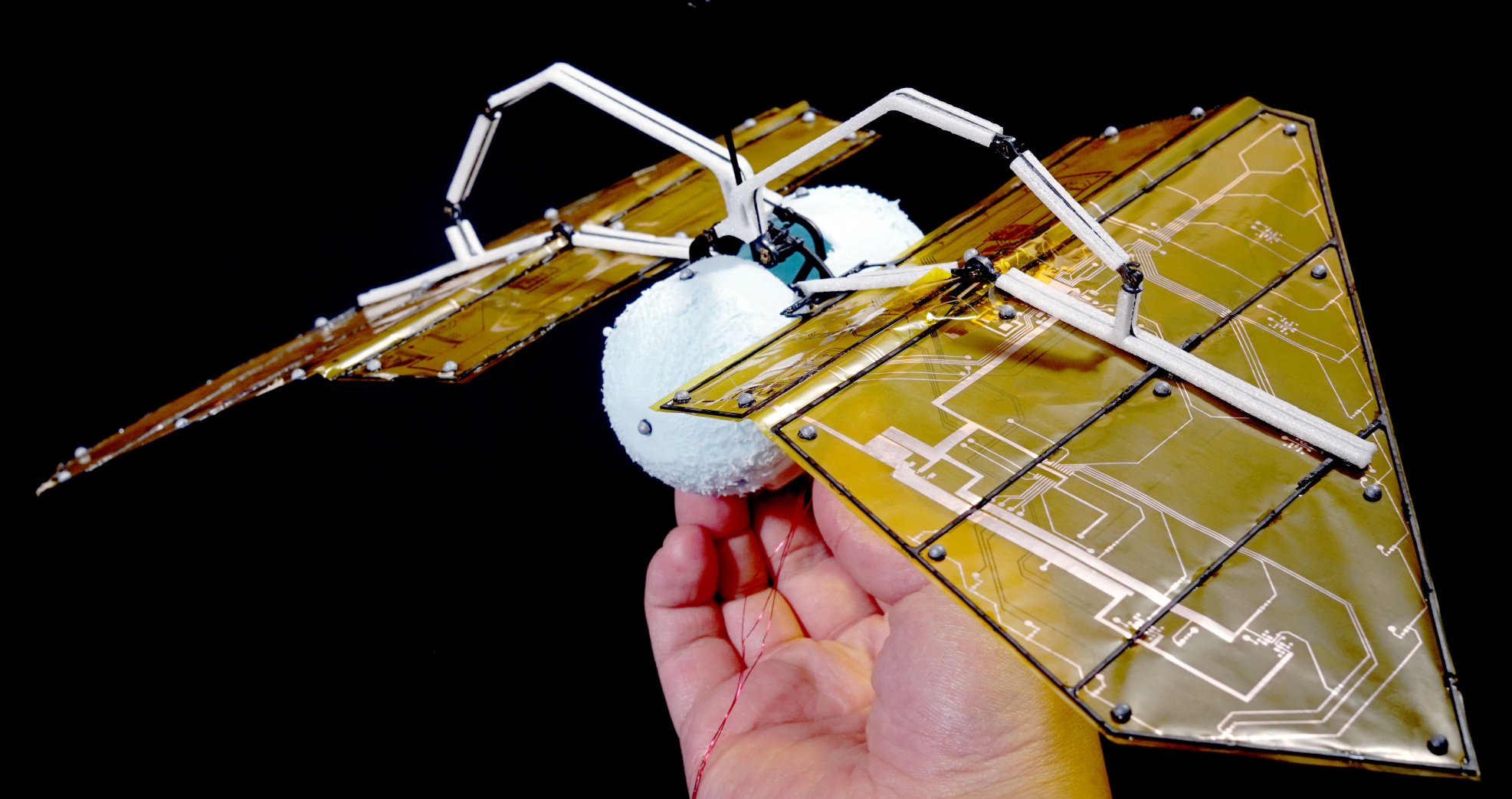}
    \caption{Northeastern University's \textit{Aerobat}, a flapping wing robot for studying dynamic morphing wing flight. This platform is designed to inspect morphology-oriented locomotion control design in MAVs. This robot captures the elbow flexion-extension which is one of the primary modes in bat flight, allowing the wing to fold and minimize negative lift during the upstroke.}
    \vspace{-0.1in}
    \label{fig:cover}
\end{figure}

% Some past flapping wing robot and why ours is unique and awesome
There are several examples of flapping wing robots, ranging from smaller insect-sized robots, or micro UAVs \cite{phan_insect-inspired_2019, farrell_helbling_review_2018, ma_controlled_2013, chukewad_robofly_2020, tu_untethered_2020, rosen_development_2016}, to bat or small bird-sized robots with a wingspan between 20 and 60 cm \cite{hoff_synergistic_2016, ramezani_bat_2016, ramezani_biomimetic_2017, hoff_optimizing_2018, sihite_computational_2020, de_croon_design_2009, peterson_wing-assisted_2011, wissa_free_2015}, and larger robots with wingspan larger than 1 m \cite{send_artificial_2012, gerdes_robo_2014}. 
% Aerobat quick introduction
We have developed a small flapping-wing robot with dynamic wing morphing inspired by bats, called the \textit{Aerobat}, which can be seen in Fig. \ref{fig:cover}. This robot features a flexible wing membrane and captures both the plunging and elbow flexion and extension which are two of the important modes in bat flapping gait \cite{sihite_computational_2020}. This robot will be described in more detail in Section \ref{sec:aerobat}. 

% Aerodynamic complexity, past simulation work, and motivation
Simulation and dynamical modeling can be extremely useful tools in studying and developing proof of concepts to further improve our robot. In particular, controlling flapping-wing robots have been an extremely challenging problem \cite{ramezani_lagrangian_2015, mangan_describing_2017, mangan_reducing_2017, hoff_trajectory_2019, sihite_enforcing_2020}. As part of our past work, we have developed a simulation model to investigate control methods \cite{sihite_enforcing_2020}, and incorporating \textit{mechanical intelligence} through a change in morphology using small, low-energy actuation to stabilize pitch dynamics \cite{sihite_integrated_2021}. In our past work, we implemented an aerodynamic model developed by Dickinson \cite{sane_lift_2001}, where they characterized the quasi-steady lift and drag coefficients of a robot based on a fruit fly. This robot does not feature a dynamically transforming wing and might not be an accurate model for our robot and does not model the leading- or trailing-edge vortex shedding \cite{hedenstrom_bat_2015, hubel_wake_2010}. This motivates us to develop a more accurate model for our simulation by incorporating unsteady aerodynamics into the model.

% Summary of this work and table of content
While it is possible to model the aerodynamic lift and drag using fluid-structure interaction between the wing surface and the air \cite{willis2007computational}, it can be very computationally expensive and difficult to formulate. Instead, we follow a simpler approach where we can use a small number of states to derive the aerodynamic forces and present them in a simplified state-space form. In this work, we derive an unsteady aerodynamic model based on Prandtl's lifting line theory and Wagner's model to more accurately model the lift generation due to the wake structure and leading-edge vortices. This model follows similar derivations to the work done by Boutet \cite{boutet_unsteady_2018} and Izraelevitch \cite{izraelevitz_state-space_2017}. In both works, they incorporate the blade elements method and vortex circulation distribution by modeling the trailing vortex as an infinitely long horseshoe vortex. We tailored these derivations to work with the Aerobat's unique morphing wing structure, which adds complexity, and present it in a simple state space form. 

Our contributions in this work are as follows: identification, clear elaboration, and validation of a theoretical approach (through simulation and experiment) that does not depend on heavy numerical computations, yet it can reasonably predict complex fluid-structure interactions. We present experimental data to support an aerodynamic model that can be integrated within a controller and can be executed in real-time for flight control and simulations of complex morphing systems, such as our Aerobat.
This paper is outlined as follows: Aerobat robot mechanical details, kinematic and dynamic modeling, aerodynamic modeling, experimental and simulation results, followed by the conclusions and future work.

% Our novel contributions in this work is as follows: (i) identification, (ii) clear elaboration, and (iii) validation of a theoretical approach (through simulation and experiment) that does not depend on heavy numerical computations, yet it can reasonably predict complex fluid-structure interactions. we present experimental data to support that an indical model with elegant form that can be integrated within a controller can be executed in real-time for flight control of complex morphing systems. Our access to NU's Aerobat, which possesses morphing wings, offered the opportunity to validate this state-space model for the first time. 

\section{Aerobat Platform}
\label{sec:aerobat}

% Aerobat introduction
This section outlines the flapping-wing UAV developed in Northeastern University called the \textit{Aerobat}, which is featured in Figure \ref{fig:cover}. This robot weighs 19.5 grams and has a wingspan of around 30 cm, and features a dynamically morphing flapping wing. The wing joints are articulated by a computational structure called the \textit{kinetic sculpture} (KS) \cite{sihite_computational_2020}, which is shown in Figure \ref{fig:ks}. The entire flapping mechanism is powered by a single motor which actuates the KS to animate the robot's flapping gait. The KS allows them flapping gait to include the elbow flexion and extension, which is one of the key modes in bat flight. This allows the wing to folds the wing during the upstroke, resulting in a reduction in a negative lift during the upstroke. This wing dynamic morphing is key to achieving a more efficient flapping gait. The robot is capable of flapping at a frequency of up to 8 Hz when powered using a 2 cell LiPo battery.

% Component Details
The wing membranes attached to the robot were fabricated with a layer of flexible PCB etched on the wing surface to carry some electronics and sensors, such as the microcontroller and IMU. The linkages forming the KS were fabricated out of carbon fiber plates which were cut using the \textit{LPKF ProtoLaser U4} UV laser cutter. This machine uses a scanner-guided laser with a wavelength of 355 nm in the UV spectrum. It is capable of producing an extremely concentrated, high-intensity beam with a laser spot size of about 20 $\mu$m. This allows us to fabricate small, lightweight, and highly detailed components for our robot. 

\section{Dynamic Modeling}
\label{sec:modeling}

\begin{figure}
    \centering
    \vspace{0.1in}
    \includegraphics[width = \linewidth]{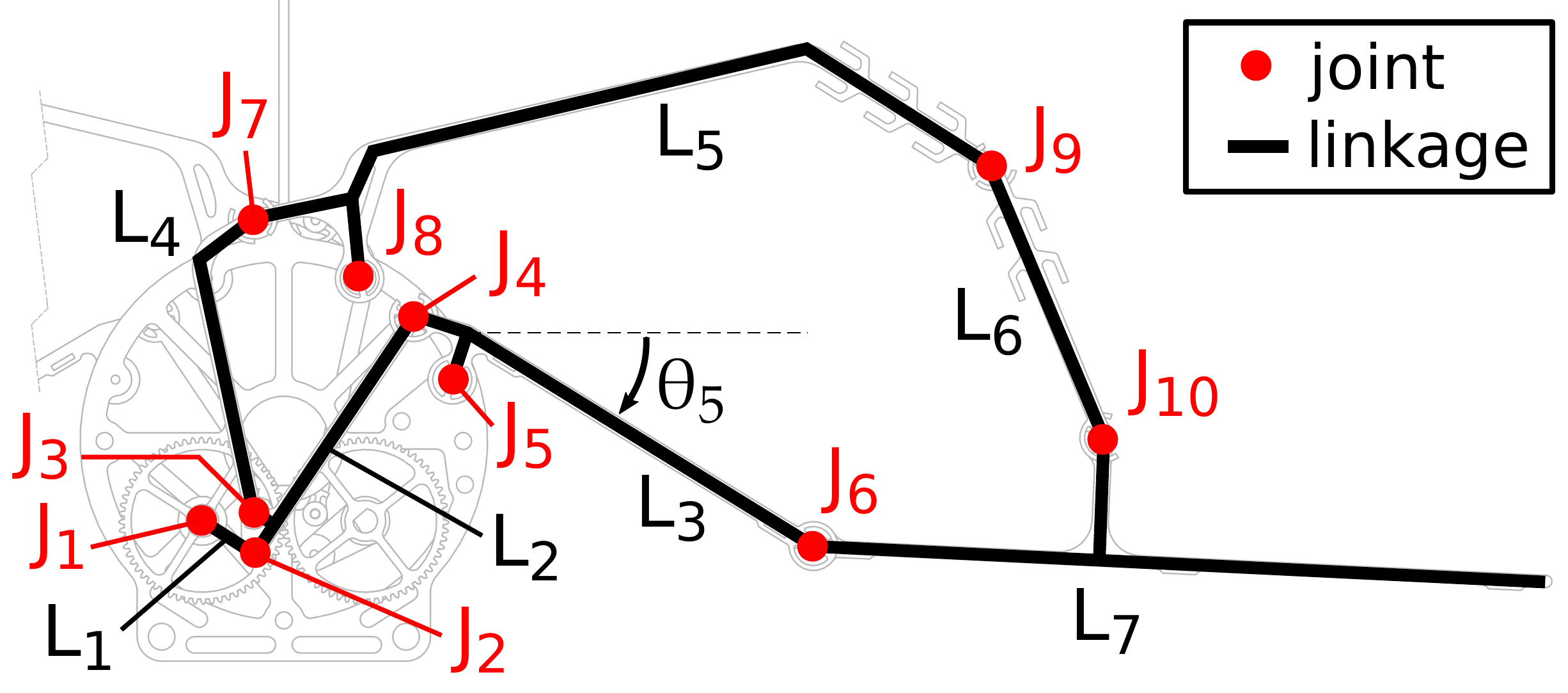}
    \caption{The kinetic sculpture of the Aerobat consists of 10 joints and 7 linkages. Joint 1 represents the gear or motor angle that drives the linkages and forms the flapping gait. Joints 5 and 6 represent the wing shoulder and elbow joint, respectively. The wings are attached to linkages 3 and 7 which represent the proximal and distal wings, respectively.}
    \vspace{-0.1in}
    \label{fig:ks}
\end{figure}

\begin{figure}
    \centering
    \vspace{0.1in}
    \includegraphics[width = \linewidth]{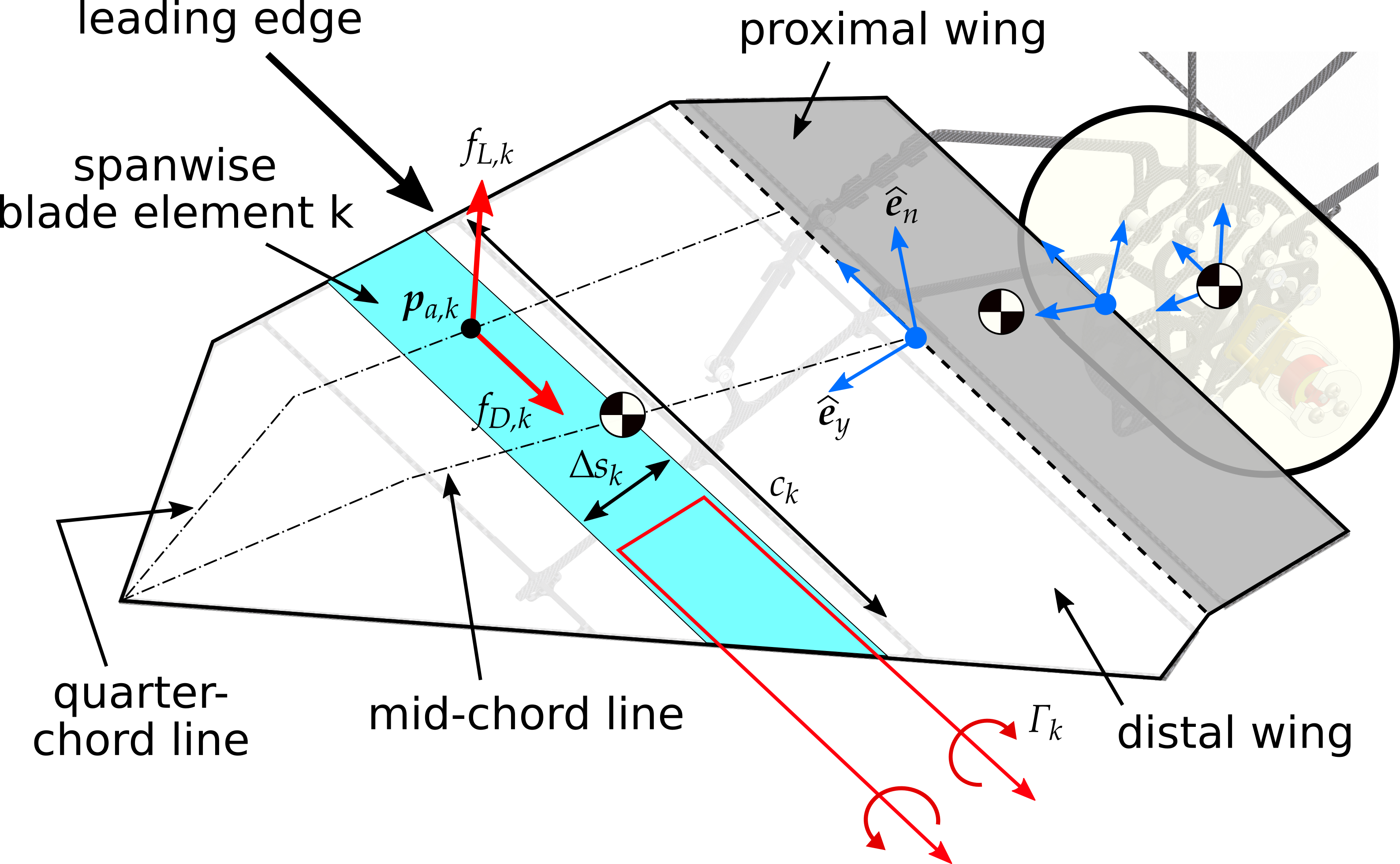}
    \caption{A diagram showing the center of masses of the system of the robot (the body and one side of the wing), in addition to the blade element diagram showing the aerodynamic forces, chord lines, and the infinite length horseshoe vortex circulation ($\Gamma_k$).}
    \vspace{-0.1in}
    \label{fig:blade_elements}
\end{figure}

% Dynamical system introduction
The Aerobat has 20 degrees-of-freedom (DOF) present in the system which makes the dynamic modeling very difficult to derive. Therefore, the following simplifications are applied to reduce the complexity of the derivations: couple the left and right wing linkages, assume massless KS linkages, and enforce KS kinematic constraints. These simplifications result in a 7 DOF system, where the KS has 1 DOF and the body has 6 DOF for the center of mass (COM) linear position and orientation. 

% Kinematic formulations
The simplified KS system can be derived as a constrained planar linkages system. Let $q_i$ and $\bm p_i \in \mathbb{R}^2$ be angle and body frame positions of joint $i \in \{1, \dots, 10\}$, respectively, as labeled in Fig. \ref{fig:ks}. Let $\bm p_i^j$ be the joint position $\bm p_i$ derived from the body fixed position $\bm p_j$. The equation of motion can be formed from the following kinematic relationship:
\begin{equation}
\begin{aligned}
    \ddot {\bm p}_4^1 &= \ddot {\bm p}_4^5, \;\;\; &
    \ddot {\bm p}_7^1 &= \ddot {\bm p}_7^8, \;\;\; &
    \ddot {\bm p}_{10}^5 &= \ddot {\bm p}_{10}^8, \;\;\;&
    \ddot q_1 &= u_k,
\end{aligned}
\label{eq:ks_constraints}
\end{equation}
where $u_k$ is the motor angular acceleration and also the only DOF of the simplified system. The constraints in \eqref{eq:ks_constraints} results in 7 equations which can be solved for the joint accelerations $\ddot{\bm q}_k = [\ddot{q}_1, \ddot{q}_2, \ddot{q}_3, \ddot{q}_5, \ddot{q}_6, \ddot{q}_8, \ddot{q}_9]^\top$. Let $\bm x_k = [\bm q_k^\top, \dot{\bm q}_k^\top]^\top$ be the states representing the kinematics equations. Then, the kinematic subsystem can be derived as follows:
\begin{equation}
\Sigma_{K}
\begin{cases}
    \dot{\bm x}_k = \bm f_k(\bm x_k) + \bm g_k(\bm x_k)\, u_k \\
    \bm y_k = [\ddot q_6, \ddot q_7]^\top = \bm C_k \, (\bm f_k(\bm x_k) + \bm g_k(\bm x_k)\, u_k),
\end{cases}
\label{eq:ks_eom}
\end{equation}
where $\bm y_k$ is the shoulder and elbow joint accelerations. $\bm y_k$ will be used as a constraint to the dynamical system equation of motion to drive the flapping gait.

% Dynamic formulations
The Aerobat system can be defined using five bodies, as shown in Fig. \ref{fig:ks}, which consists of the main body, proximal and distal wing segments of both wings. Let the superscript $B$ represents the vector defined in body frame, e.g., $\bm z^{B} \in \mathbb{R}^3$. The transformation from body to inertial frame is defined using the rotation matrix $\bm R_B$, i.e., $\bm z = \bm R_B\, \bm z^B$. The dynamical equation of motion can then be derived using Euler-Lagrangian dynamic formulations, using a similar methodology as our previous work in \cite{sihite_enforcing_2020, sihite_integrated_2021}. 

Let $\bm q_d = [\bm p_B^\top, q_s, q_e]^\top$ be states representing the body inertial center of mass position ($\bm p_B \in \mathbb{R}^3$), shoulder and elbow joint angle ($q_s$ and $q_e$, respectively). Furthermore, let $\bm \omega_B \in \mathbb{R}^3$ be the angular speed of the body, also defined as $\dot{\bm R}_B = \bm R_B [\bm \omega_B]_\times$, where $[\,\cdot\,]_\times$ is the skew-symmetric operator. Finally, let $\bm a_d = [\ddot{\bm q}_d^\top, \dot{\bm \omega}_B^\top]^\top$ be the acceleration vector of the dynamical states. Solving the Euler-Lagrangian equation of motion derivations and applying the kinematic constraints to the wing joints results in the following constrained equation of motion:
\begin{align}
    \bm M_d(\bm q_d, \bm R_B) \, \bm a_d &= \bm h_d(\bm q_d, \dot{\bm q}_d, \bm R_B, \bm \omega_B) + \bm u_a + \bm J_c^\top \bm \lambda \label{eq:dynamic_eom_base} \\
    \bm J_c \, \bm a_d &= [\ddot q_s, \ddot q_e]^\top = \bm y_k, \label{eq:dynamic_constraints}
\end{align}
where $\bm M_d$ is the mass and inertia matrix, $\bm h_d$ is a vector containing the coriolis and gravitational terms, $\bm u_a$ is the combined aerodynamic generalized forces, and $\bm J_c^\top \bm \lambda$ is the forces to enforce the constraints in \eqref{eq:dynamic_constraints}. Both \eqref{eq:dynamic_eom_base} and \eqref{eq:dynamic_constraints} can be solved algebraically for $\bm a_B$ and $\bm \lambda$. Let $\bm x_d = [\bm q_d^\top, \dot{\bm q}_d^\top, \bm r_B^\top, \bm \omega_B]$, where $\bm r_B$ is the elements of $\bm R_B$ truncated into a vector form. Then, we can derive the following dynamical subsystem:
\begin{equation}
\Sigma_{D}:\quad 
    \dot{\bm x}_d = \bm f_d(\bm x_d) + \bm g_{d,1}(\bm x_d)\, \bm u_a + \bm g_{d,2}(\bm x_d) \, \bm y_k.
\label{eq:dynamic_eom}
\end{equation}
The aerodynamic forces ($\bm u_a$) acting on the system are derived in the following section.

\section{Aerodynamic Modeling}
\label{sec:aerodynamic_modeling}

% introduction on the aerodynamics modeling
Modeling the aerodynamics of a flapping wing can be very complex and computationally expensive, since the lift generated depends on the fluid-structure interaction between the wing and the air. An unsteady aerodynamic model, such as \cite{boutet_unsteady_2018} and \cite{izraelevitz_state-space_2017}, can be difficult to implement due to the very high angle of attack present in our design, which depending on the airflow can be more than 50$^\circ$. Here, we can use a simple quasi-steady blade element model for its simplicity using a similar method to our previously developed simulation model in \cite{sihite_enforcing_2020}. 

\subsection{Quasi-steady aerodynamics modeling}

The aerodynamic forces can be modeled by following blade element theory which is done by segmenting the wing surface in several discrete blade elements and evaluating the aerodynamic forces locally. The Aerobat's morphing wing can be categorized into four wing segments: proximal and distal wing segments on each left and right wing. To simplify the problem, we represent all wing segments as continuous, unsegmented blade elements for the lifting line theory calculations. Additionally, we assume that the wing surface is rigid which significantly reduces the modeling complexity.

Fig.~\ref{fig:blade_elements} illustrates the chordwise blade element of the distal wing segment, where the wing chord length varies on each blade element $k$. Let $\bm p_{a,k}$ be the blade element's aerodynamic center of pressure of blade element $k$ which is located at a quarter-chord distance from the leading edge. Additionally, $\bm p_{a,k}$ is defined about the inertial frame. Then we can map the forces acting on $\bm p_{a,k}$ by solving for the virtual displacement:
\begin{equation}
\begin{aligned}
    \bm u_{a,k} = \bm B_k \, \bm f_{a,k}, 
    \qquad
    \bm B_k = \left( \frac{\partial \dot{\bm p}_{a,k}}{\partial \bm v_d} \right)^\top,
\end{aligned}
\label{eq:virtual_disp}
\end{equation}
where $\bm f_{a,k}$ represents the blade element's aerodynamic force in inertial frame, and $\bm v_d = [\dot{\bm q}_d^\top, \bm \omega_B^\top]^\top$ is the generalized coordinate velocities. Here, $\bm f_{a,k}$ is defined as follows:
\begin{equation}
\begin{aligned}
    f_{L,k} &= \tfrac{1}{2} \rho |\bm v_{k}|^2 \, C_L(\alpha_k) \, c_k \, \Delta s_k\\
    f_{D,k} &= \tfrac{1}{2} \rho |\bm v_{k}|^2 \, C_D(\alpha_k) \, c_k \, \Delta s_k \\
    \bm f_{a,k} &= f_{L,k}\, \hat{\bm e}_{L,k} + f_{D,k}\, \hat{\bm e}_{D,k},
\end{aligned}
\label{eq:strip_aero_force}
\end{equation}
where $\rho$ is the air density, $c_k$, $\Delta s_k$, and $\alpha_k$ are the chord length, span width, and angle of attack of the blade element $k$, respectively. $\bm v_{k}$ in \eqref{eq:strip_aero_force} is defined as the effective airspeed along the directions $\hat {\bm e}_{y}$ and $\hat {\bm e}_{n}$ (along the chord and normal to the wing surface), as illustrated in Fig.~\ref{fig:blade_elements}. The unit vectors $\hat{\bm e}_{L,k}$ and $\hat{\bm e}_{D,k}$ are the lift and drag forces directions of the blade element $k$, respectively. $\hat{\bm e}_{D,k}$ is parallel to $\bm v_k$, while $\hat{\bm e}_{L,k}$ is perpendicular to $\bm v_k$ and facing the top side of the wing. $C_L$ and $C_D$ are the lift and drag coefficients, which are typically acquired experimentally. In this work, we use the lift and drag coefficients formulated by Dickinson \cite{sane_lift_2001}, as follows:
\begin{equation}
\begin{aligned}
    C_L(\alpha) &= 0.225 + 1.58 \, \sin(2.13 \, \alpha - 7.2^\circ)\\
    C_D(\alpha) &= 1.92 - 1.55 \, \cos(2.04 \, \alpha - 9.82^\circ),
\end{aligned}    
\label{eq:dickinson_model}
\end{equation}
where $\alpha$ is in degrees. Finally, the combined generalized forces $\bm u_a = \sum_{k \in \mathcal{W}} \bm u_{a,k}$ can then be calculated by adding up all of the generalized aerodynamics forces defined by the set $\mathcal{W}$ which contains all blade elements.

\subsection{Unsteady lift aerodynamic model using Wagner's function}
\label{subsec:unsteady_aero}

The unsteady lift aerodynamic model derived in this section follows similar derivations to \cite{boutet_unsteady_2018}. This model uses the lifting line theory and Wagner's function to develop a model for calculating the lift coefficient. Let $S$ be the total wingspan and $y \in [-S/2, S/2]$ represents a position along the wingspan. The circulation distribution on the wing can be defined as a function of truncated Fourier series of size $m$ across the wingspan, as follows:
\begin{equation}
\begin{gathered}
    \Gamma(t,y) = \frac{1}{2} a_0 \, c_0 \, U \, \sum^{m}_{n=1} a_n(t) \, \sin(n\,\theta(y))
\end{gathered}
\end{equation}
where $a_n$ is the Fourier coefficients, $a_0$ is the slope of the angle of attack, $c_0$ is the chord length at wing's axis of symmetry, and $U$ is the free stream airspeed. Then $\theta$ is the change of variable defined by $y = (S/2)\cos(\theta)$ for describing a position along the wingspan $y \in (-S/2, S/2)$. From $\Gamma(t,y)$, we can derive the additional downwash induced by the vortices, defined as follows:
\begin{equation}
\begin{aligned}
    w_{y}(t,y) & 
    = -\frac{1}{4\pi} \int_{-S/2}^{S/2} \frac{d \Gamma / d y_0}{y - y_0} dy_0 \\ &
    = - \frac{a_0 c_0 U}{4S} \sum^{m}_{n=1} n a_n(t)  \frac{\sin(n \theta)}{\sin(\theta)}.
\end{aligned}
\label{eq:induced_downwash}
\end{equation}
%
%Note that \eqref{eq:induced_downwash} has a division by $\sin(\theta)$, which means that this equation is singular for $\theta \in \{0, \pm \pi\}$, or the $\theta$ values representing the wingtip positions.

Following the unsteady Kutta-Joukowski theorem, the sectional lift coefficient can be expressed as follows:
\begin{equation}
\begin{aligned}
    C_L(t,y) &= \frac{2 \Gamma}{U c(y)} + \frac{2\dot{\Gamma}}{U^2} \\
        &= a_0 \sum^{m}_{n=1} \left( \frac{c_0}{c(y)} a_n(t) + \frac{c_0}{U} \dot{a}_n(t) \right) \sin(n\theta),
\end{aligned}
\label{eq:lift_coeff_fourier}
\end{equation}
where $c(y)$ is the chord length at the wingspan position $y$. The computation of the sectional lift coefficient response of an airfoil undergoing a step change in downwash $\Delta w(y) \ll U$ can be expressed using Wagner function $\Phi(t)$:
\begin{equation}
\begin{aligned}
    c_L(t,y) &= \frac{a_0}{U} \Delta w(t,y) \Phi(\tilde t) \\
    \Phi(\tilde t)    &= 1 - \psi_1 e^{-\epsilon_1 \tilde t} - \psi_2 e^{-\epsilon_2 \tilde t}
\end{aligned}
\label{eq:lift_coeff_wagner}
\end{equation}
where $\tilde t(t) = \int_0^t (v_e^i/b) dt$ is the normalized time which is defined as the distance traveled divided by half chord length ($b = c/2$). Here, $v_e^i$ is defined as the velocity of the quarter chord distance from the leading edge in the direction perpendicular to the wing sweep. For the condition where the freestream airflow dominates $v_e$, then we can approximate the normalized time as $\tilde t = Ut/b$. The Wagner model in \eqref{eq:lift_coeff_wagner} uses Jones' approximation \cite{boutet_unsteady_2018}, with the following coefficients: $\psi_1 = 0.165$, $\psi_2 = 0.335$, $\epsilon_1 = 0.0455$, and $\epsilon_2 = 0.3$.

\begin{figure}[t]
    \centering
    \vspace{0.1in}
    \includegraphics[width = 0.75\linewidth]{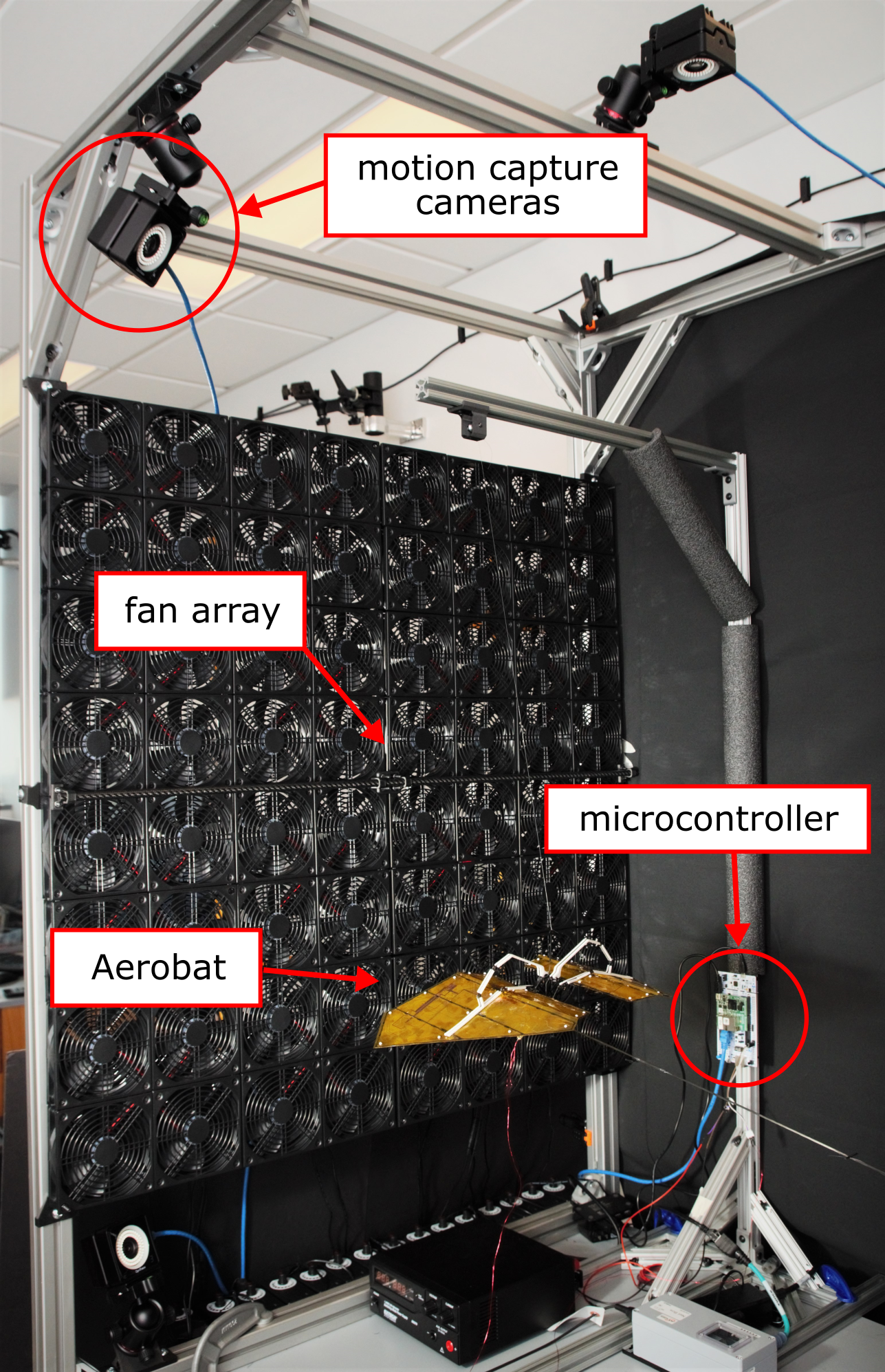}
    \caption{The flight test arena in Northeastern University for studying flapping-wing robots. This arena is equipped with fan arrays, load cell, motion capture, and a high-speed camera. The fan arrays generate a steady and uniform airflow towards the robot as the robot is flapping on the load cell to capture the generated forces and moments.}
    \vspace{-0.1in}
    \label{fig:rise}
\end{figure}

\begin{comment}
\begin{figure}[t]
    \centering
    \vspace{0.1in}
    \includegraphics[width = \linewidth]{figures/plot_wagner_vs_dickinson.pdf}
    \caption{Load cell experimental data vs. the simulated lift and drag generated by both models derived in Section \ref{sec:aerodynamic_modeling} under the same airflow and flapping conditions (1.65 m/s airspeed and 4.5 Hz flapping frequency).}
    \vspace{-0.1in}
    \label{fig:results}
\end{figure}
\end{comment}

\begin{figure*}[t]
    \centering
    \vspace{0.1in}
    \hspace{0.25in}
    \textbf{Slow Flapping Speed}
    \hspace{2.25in}
    \textbf{Fast Flapping Speed}
    \\
    \includegraphics[width = 0.48\linewidth]{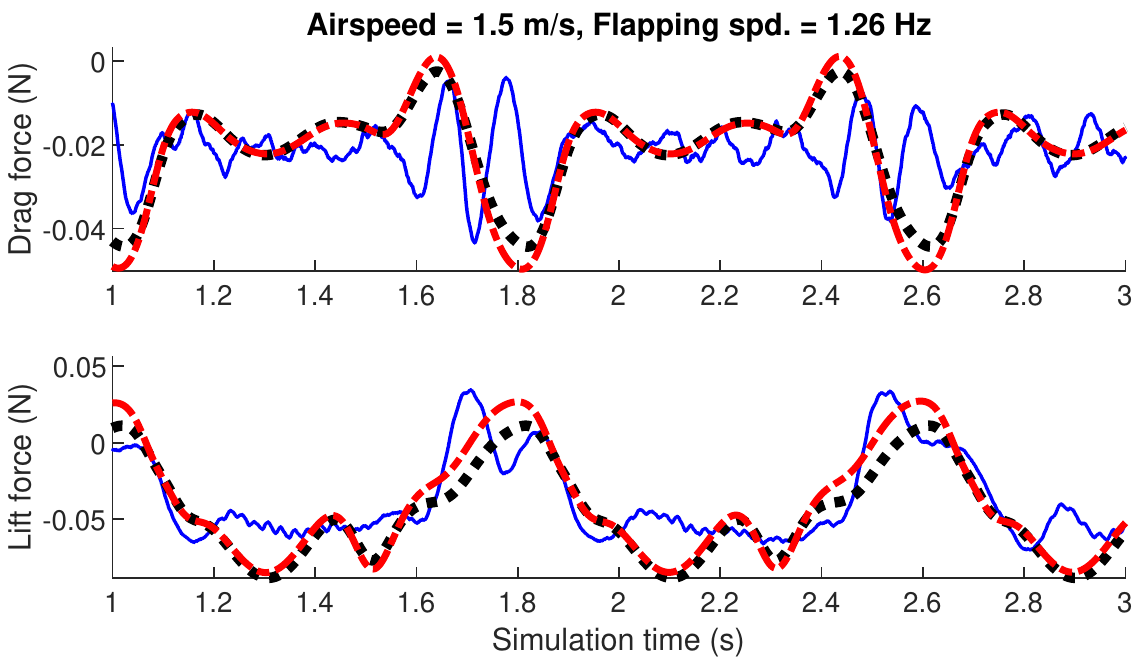}
    \hfill
    \includegraphics[width = 0.48\linewidth]{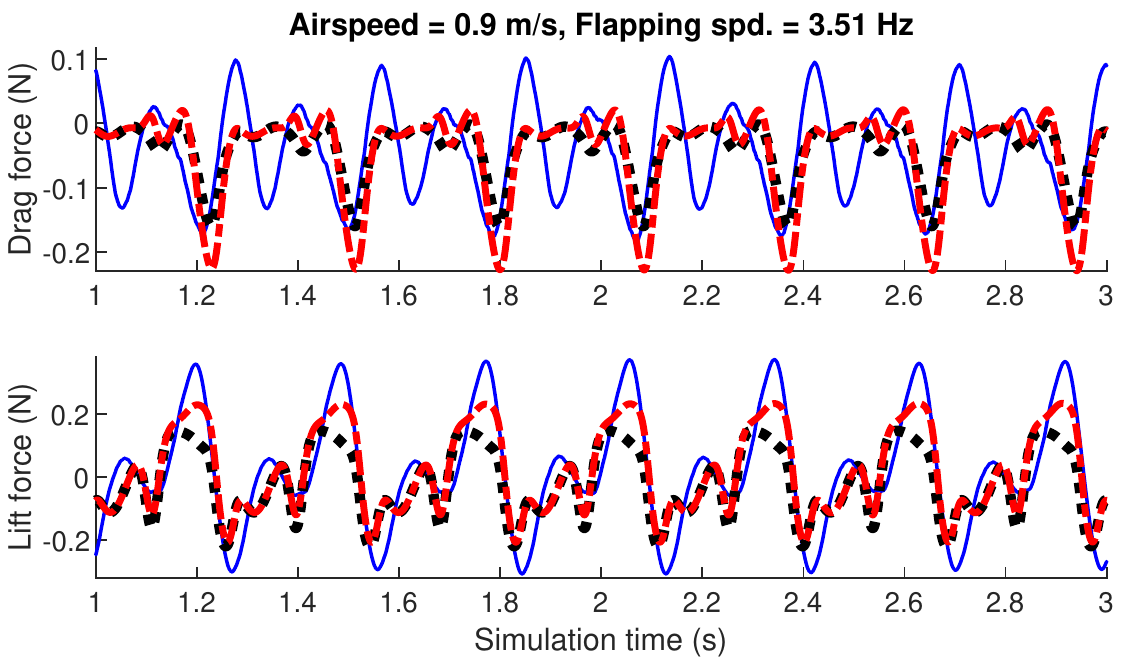}
    \\ \vspace{0.1in}
    \includegraphics[width = 0.48\linewidth]{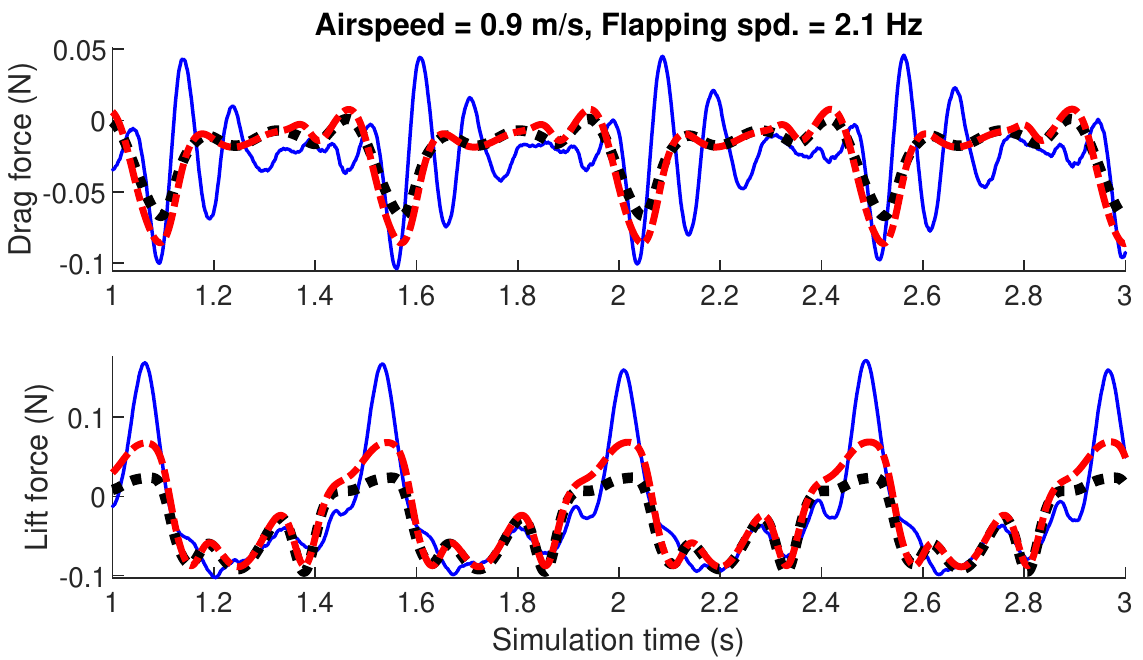}
    \hfill
    \includegraphics[width = 0.48\linewidth]{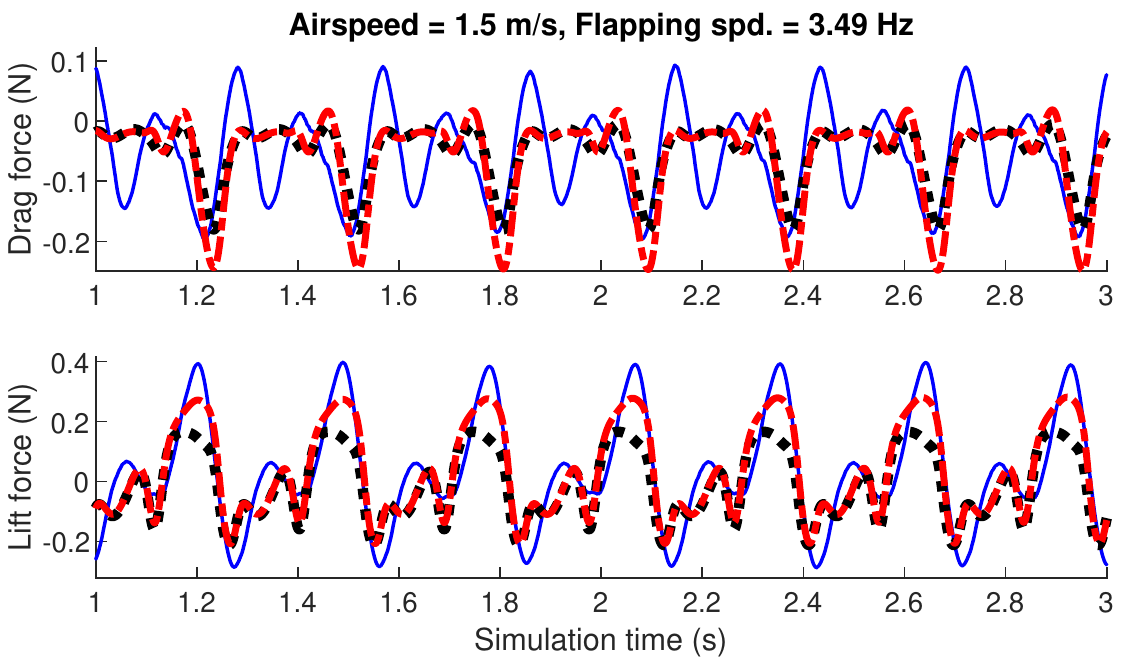}
    \\ \vspace{0.1in}
    \includegraphics[width = 0.48\linewidth]{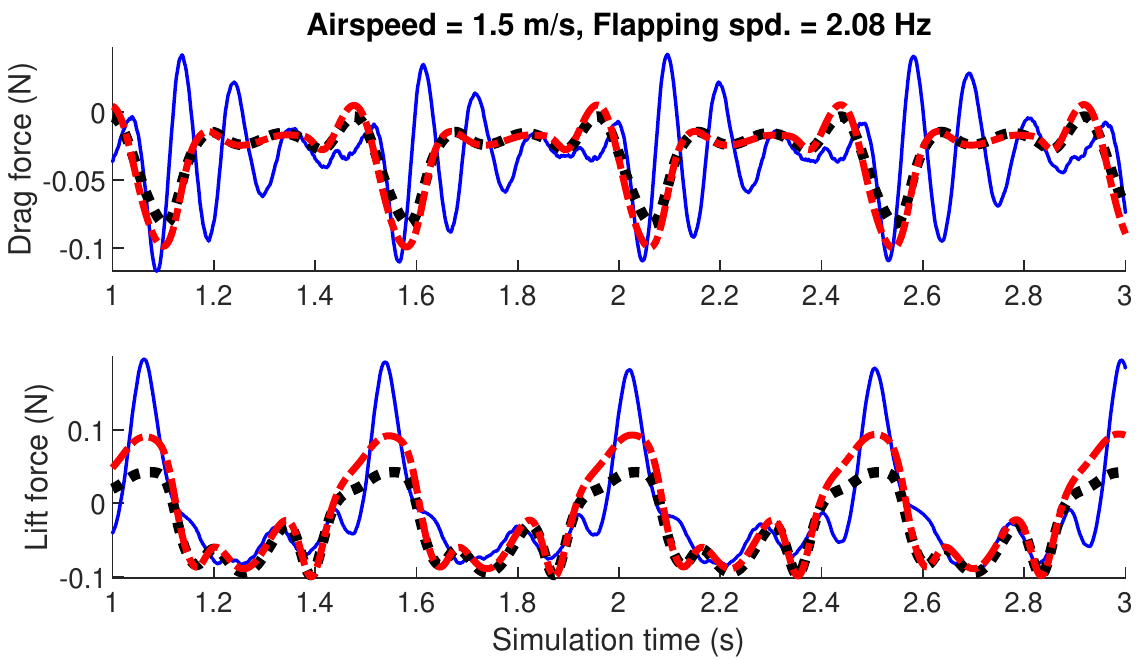}
    \hfill
    \includegraphics[width = 0.48\linewidth]{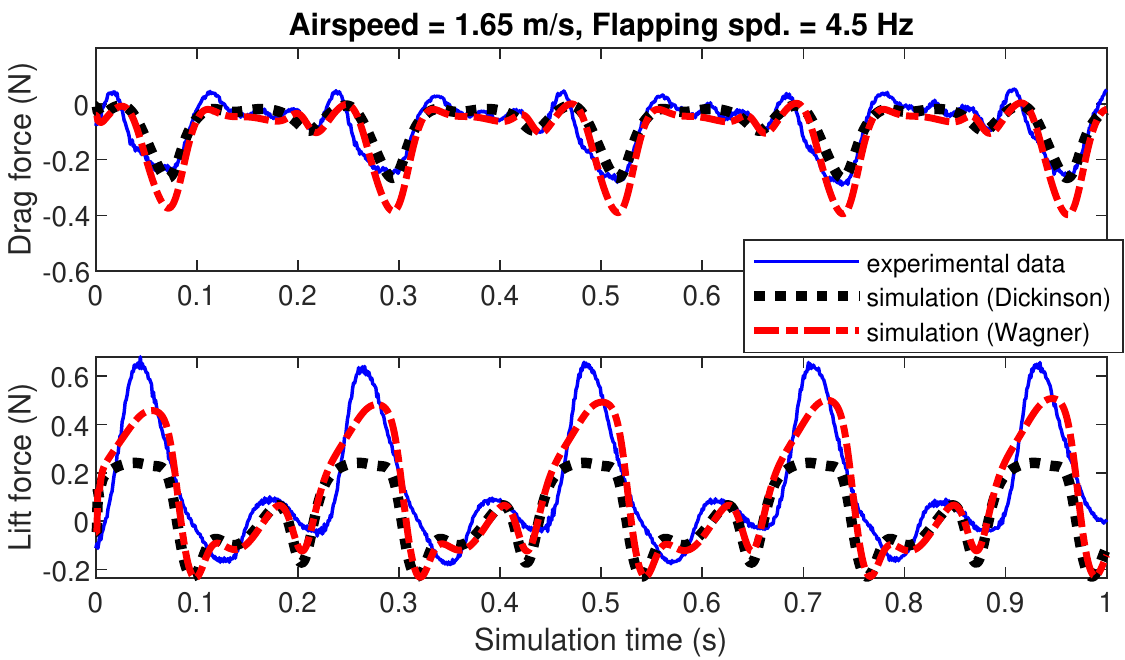}
    \caption{Load cell experimental data (solid line) vs. the simulated lift and drag generated by the quasi-steady Dickinson's model (dotted line) and Wagner aerodynamic model (dashed line). The robot was subjected to various airspeed and flapping frequencies, which was then simulated using the model derived in Section \ref{sec:aerodynamic_modeling} under the same conditions.}
    \vspace{-0.1in}
    \label{fig:results}
\end{figure*}

\begin{figure*}[t]
    \centering
    \vspace{0.1in}
    \includegraphics[width = \linewidth]{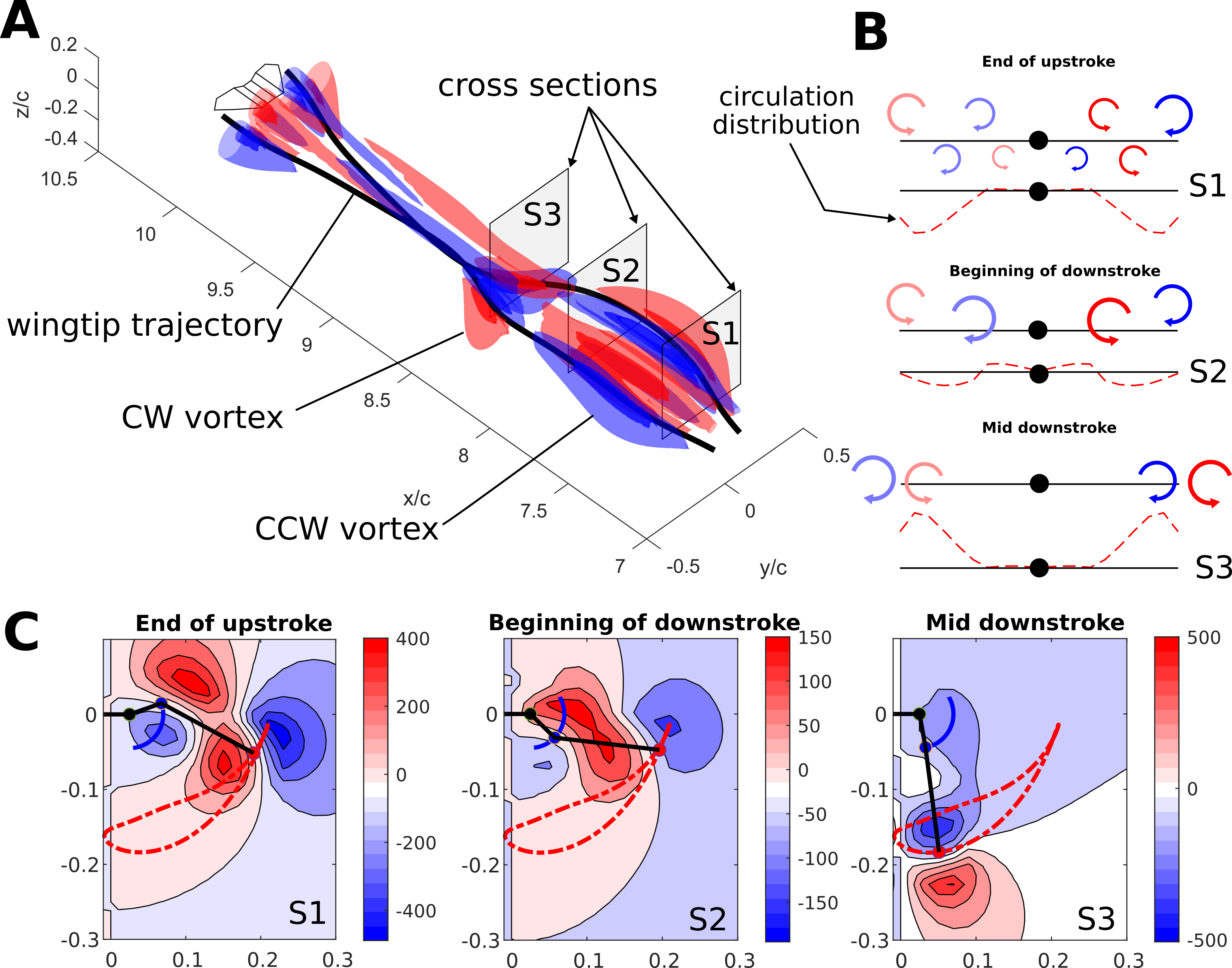}
    \caption{Simulation result of the vortex wake within a flapping period. The color red and blue indicate counter-clockwise and clockwise vortex along the robot's front axis, respectively. \textbf{(A)} The propagated induced vortex over a single flapping period, in addition to the cross-sectional slice indicating the end of upstroke, beginning of downstroke, and mid downstroke timings. \textbf{(B)} The circulation distribution and induced vortices across the wingspan at the specified timings, as seen from the wing's front side. \textbf{(C)} The contour plot of the induced vortex distribution at the specified timings, as seen from the wing's front side.}
    \vspace{-0.1in}
    \label{fig:vortex_plot}
\end{figure*}

%\begin{figure*}[t]
    %\centering
    %\vspace{0.1in}
    %\includegraphics[width = \linewidth]{figures/plot_vortex_oscillate.png}
    %\caption{Simulation result of the vortex wake within a flapping period, as the robot pitches oscillates up and down.}
%    \vspace{-0.1in}
%    \label{fig:vortex_plot_pitch}
%\end{figure*}

Duhamel's principles can be used to superimpose the transient response due to a step change in downwash as defined in \eqref{eq:lift_coeff_wagner}. Additionally, integration by parts can be used to simplify the equation further, resulting in the following equation:
\begin{equation}
\begin{aligned}
    C_L(t,y) &= \frac{a_0}{U} \left( w(t,y) \Phi(0) - \int_{0}^{t} \frac{\partial \Phi(t - \tau)}{\partial \tau} w(\tau, y) d\tau \right).
\end{aligned}
\label{eq:aero_CL_base}
\end{equation}
\begin{equation}
\begin{aligned}
    \frac{\partial \Phi(t - \tau)}{\partial \tau} &=
    -\frac{\psi_1 \epsilon_1 U}{b} e^{-\frac{\epsilon_1 U}{b}(t-\tau)}
    -\frac{\psi_2 \epsilon_2 U}{b} e^{-\frac{\epsilon_2 U}{b}(t-\tau)}
\end{aligned}
\label{eq:partial_phi}
\end{equation}
Here, $w(t,y)$ is the total downwash defined as:
\begin{equation}
    w(t,y) = v_n(t,y) + w_y(t,y),
\label{eq:total_downwash}
\end{equation}
where $v_n$ is the airfoil velocity normal to the wing surface which depends on the freestream velocity and the inertial dynamics. Finally, we can represent the integrals as the following states:
\begin{equation}
\begin{aligned}
    z_{1} (t,y) &= \int_{0}^{t} \frac{\psi_1 \epsilon_1 U}{b} e^{-\frac{\epsilon_1 U}{b}(t-\tau)} w(\tau,y) d\tau
    \\
    z_{2} (t,y) &= \int_{0}^{t} \frac{\psi_2 \epsilon_2 U}{b} e^{-\frac{\epsilon_2 
    U}{b}(t-\tau)} w(\tau,y) d\tau.
\end{aligned}
\label{eq:aero_states_z}
\end{equation}
Both of these states can be expressed as an ODE by deriving the time derivatives of \eqref{eq:aero_states_z}. They can be derived using Leibniz integral rule, yielding the following equations:
\begin{equation}
\begin{aligned}
    \dot z_{1} (t,y) &= \frac{\psi_1 \epsilon_1 U}{b} \left( w(t,y) - \frac{\epsilon_1 U}{b} z_1(t,y) \right) \\
    \dot z_{2} (t,y) &= \frac{\psi_2 \epsilon_2 U}{b} \left( w(t,y) - \frac{\epsilon_2 U}{b} z_2(t,y) \right).
\end{aligned}
\label{eq:aero_states_dz}
\end{equation}
The sectional lift coefficient can then be defined as:
\begin{equation}
\begin{aligned}
    C_L(t,y)  = \frac{a_0}{U} \left( w(t,y) \phi(0) + z_1(t,y) + z_2(t,y) \right),
\end{aligned}
\label{eq:aero_CL_final}
\end{equation}
and we can march the aerodynamic states $z_1$ and $z_2$ forward in time using \eqref{eq:aero_states_dz}. 

Finally, we can relate the both sectional lift coefficient equations in \eqref{eq:lift_coeff_fourier} and \eqref{eq:aero_CL_final} to solve for the Fourier coefficient rate of change, $\dot{a}_n$. The aerodynamic states are defined along the span of the wing and we can discretize them into $m$ blade elements. Therefore, we can derive the $m$ equations relating \eqref{eq:lift_coeff_fourier} and \eqref{eq:aero_CL_final} on each blade element to solve for the $\dot{a}_n$. Then, including $z_1$ and $z_2$ on each blade elements, we will have $3m$ ODE equations to solve. We can represent $a_n$, $z_1$, and $z_2$ of all blade elements as the vector $\bm a_n \in \mathbb{R}^{m}$, $\bm {z}_1 \in \mathbb{R}^{m}$, and $\bm z_2 \in \mathbb{R}^{m}$, respectively. Only the lift coefficient is derived in this model. Therefore, we use the quasi-steady drag coefficient in \eqref{eq:dickinson_model} in the simulation.

\subsection{State space representation}

Let the vector $\bm \zeta = [\bm a_n^\top, \bm z_1^\top, \bm z_2^\top] \in \mathbb{R}^{3m}$ contains all unsteady aerodynamic states defined in Section \ref{subsec:unsteady_aero}. The effective downwash $w(t,y)$ in \eqref{eq:total_downwash} can be expressed as:
\begin{equation}
\begin{aligned}
    w(t,y) = \bm f_w(\bm x_d(t), y) + \bm g_w(y)^\top \bm \zeta(t).
\end{aligned}
\label{eq:total_downwash_2}
\end{equation}
Then we can use \eqref{eq:total_downwash_2} to rewrite the aerodynamic states ODE and $\bm u_a$ as follows:
\begin{equation}
\Sigma_{A}
\begin{cases}
    \dot{\bm \zeta} = \bm f_\zeta(\bm x_d) + \bm g_{\zeta}\, \bm \zeta, \\
    \bm u_a = \bm h_{a1} (\bm x_d) + \bm h_{a2} (\bm x_d) \bm \zeta,
\end{cases}
\label{eq:eom_aerodynamic}
\end{equation}
which defines the unsteady aerodynamic states subsystem $\Sigma_A$. 
We can then combine all of the subsystems into the following state space form:
\begin{equation}
\Sigma_{S}
\begin{cases}
    \dot{\bm x}_k &= \bm f_k(\bm x_k) + \bm g_{k}(\bm x_k) u_k \\
    \dot{\bm x}_d &= \bm f_d(\bm x_d) + \bm g_{d,1}(\bm x_d)\, \bm u_a + \bm g_{d,2}(\bm x_d)\, \bm y_k  \\
    \dot{\bm \zeta} &= \bm f_\zeta(\bm x_d) + \bm g_{\zeta}\, \bm \zeta \\
    \bm y_k &= \bm C_k \, ( \bm f_k(\bm x_k) + \bm g_{k}(\bm x_k) u_k ) \\
    \bm u_a &= \bm h_{a,1} (\bm x_d) + \bm h_{a,2} (\bm x_d) \bm \zeta,
\end{cases}
\label{eq:eom_statespace}
\end{equation}
By defining the combined states of $\bm x = [\bm x_k^\top, \bm x_d^\top, \bm \zeta^\top]^\top$, we can rewrite \eqref{eq:eom_statespace} into the concise form:
\begin{equation}
    \dot {\bm x} = \bm f(\bm x) + \bm g(\bm x) \, u_k,
\label{eq:eom_combined}
\end{equation}
where the motor acceleration ($u_k$) is the only input to the system. Other external forces, such as the aerodynamic forces, are state-dependent and can be adjusted by changing the flapping speed with $u_k$.

\section{Simulation and Experimental Results}
\label{sec:results}

% experimental hardware
The model derived in Section \ref{sec:modeling} and \ref{sec:aerodynamic_modeling} will be compared to Aerobat's experimental data to verify the accuracy of the model. The experiment was conducted in our flight arena shown in Fig. \ref{fig:rise}. This arena features a 8 by 8 array of fans to generate a uniform airflow inside the cage, up to a maximum airspeed of 2 m/s. The uniformity of the airflow was verified using anemometer to measure the airflow at various positions in front of the fan array. The robot is then mounted on a load cell to measure the forces and moments generated by the flapping wings. We choose the \textit{ATI Nano17} as our load cell, which is calibrated for a sensing range of 12 N for forces in $x$ and $y$ directions, 17 N for force in $z$ direction, and 120 N-mm for torques. It has a fine-grained resolution of 1/320 N and 1/64 N-mm for forces and torques, respectively. This load cell uses a high-performance data acquisition unit that is capable of reaching sample rates up to 7000 Hz. Note that the robot's flapping speed can be determined from the load cell measurements. 

% Simulation set up
We captured a set of load cell measurements at various airspeed and flapping frequencies and simulated it under the same conditions. The simulation was performed in Matlab where the ODE in \eqref{eq:eom_combined} was marched forward in time using the 4'th order Runge-Kutta algorithm. Finally, we compared the simulated lift and drag forces with the load cell measurements (the load cell $x$ and $z$-axis force measurements, respectively) to verify the accuracy of our model. 

% Results discussions
Figure \ref{fig:results} shows the comparison between the simulated aerodynamic forces and the load cell measurements at six different flapping and airspeed conditions. The RMS errors between the models and measurements vary depending on the testing conditions. At the fastest airspeed of 1.65 m/s and flapping frequency of 4.5 Hz, the simulated lift and drag using the quasi-steady Dickinson's model have RMS errors of 0.180 N and 0.044 N, respectively. On the other hand, the simulated lift and drag using Wagner model have RMS errors of 0.118 N and 0.058 N, respectively. 

As shown in Fig. \ref{fig:results}, the simulated lift using the Wagner model is relatively close to the experimental data. The lift forces estimated by the quasi-steady model are noticeably less accurate compared to the Wagner model as the flapping rate increases and unsteady flow conditions appear. This result demonstrates that the proposed model is more suited for the control design and simulation of flapping-wing robot with unsteady flow, such as Aerobat. 

However, there are some oscillations present in the drag force measurements which are not present in our simulated forces, and the peaks of the simulated lift forces sometimes don't match well compared to the measurements. The errors in the simulation vs. experimental data could be caused by the rigid wing assumption used to derive the model, which ignores the structural flexibility of our robot. The wing surfaces of the Aerobat can flex and deform in response to aerodynamic loads, which may contribute to the discrepancy in the model.

% Vortex plots discussion
The flapping wing produces wake structures trailing behind the wing. We can visualize this by evaluating the vortex-induced velocities at a fixed position behind the wing's trailing edge. The induced velocities can be evaluated using Biot-Savart theorem \cite{parslew2013theoretical}. Figure \ref{fig:vortex_plot} shows the visualization of the simulated vortex-induced wake and the circulation distribution generated by the flapping wings within one gait cycle. Figures \ref{fig:vortex_plot}B and \ref{fig:vortex_plot}C show the circulation distributions along the wing surface and vortex intensity at the cross-sections shown in Fig. \ref{fig:vortex_plot}A, respectively. There is a notably very small wake intensity during the upstroke, showing the reduced negative lift generated during the upstroke as the wing folds. The experimental validation of the wake structures will be a part of our future work, which can be done by using particle image velocimetry.

\section{Conclusions and Future Work}
\label{sec:conclusion}

% summary
We have derived the models for control analysis and simulation for our robot, Aerobat, which implements the unsteady aerodynamic model based on Prandtl's unsteady lifting line theory and Wagner's function. The equation of motion is then simplified into the state space form was then simulated under similar conditions to the experiments to compare the lift and drag forces produced by the robot. The simulated lift and drag forces using the Wagner model are more closely match our experimental results compared to the quasi-steady model. The circulation distributions calculated during the simulation can be used to visualize and characterize the wake structures produced by the flapping wing.

% future work
For our future work, we will work towards validating the modeled wake structures and extend the modeling to include control inputs other than the flapping speed to stabilize the robot during flight. A more accurate aerodynamic model done in this work can be used to develop some proof of concepts of efficient actuation and control methods that we can later implement in the actual robot. We can explore the concept of embodiment and mechanical intelligence \cite{hauser_role_2012}, where the morphology and compliant structures play a key role in the feedback control. We seek to implement a minimum number of high-powered actuators such as the motor driving the KS, and use several low-power actuators using inexpensive computational resources to generate dynamic gaits instead. We can also pursue a wake-based design approach, where we iterate our design to match a target wake distribution. Here, we utilize efficient and improvable models for wake-based gait description. These models can be evaluated by an optimizer thousands of times within a fraction of a second for control and morphology design purposes.

\addtolength{\textheight}{-13cm}   % This command serves to balance the column lengths

\printbibliography

@article{sane_lift_2001,
	title = {Lift and drag production by a flapping wing},
	language = {en},
	author = {Sane, S P and Dickinson, M H},
	year = {2001},
	pages = {20}
}

@article{de_croon_design_2009,
	title = {Design, {Aerodynamics}, and {Vision}-{Based} {Control} of the {DelFly}},
	volume = {1},
	issn = {1756-8293, 1756-8307},
	url = {http://journals.sagepub.com/doi/10.1260/175682909789498288},
	doi = {10.1260/175682909789498288},
	language = {en},
	number = {2},
	urldate = {2021-03-01},
	journal = {International Journal of Micro Air Vehicles},
	author = {de Croon, G.C.H.E. and de Clercq, K.M.E. and Ruijsink, R. and Remes, B. and de Wagter, C.},
	month = jun,
	year = {2009},
	pages = {71--97}
}

@inproceedings{rosen_development_2016,
	address = {Stockholm, Sweden},
	title = {Development of a 3.2g untethered flapping-wing platform for flight energetics and control experiments},
	isbn = {978-1-4673-8026-3},
	url = {http://ieeexplore.ieee.org/document/7487492/},
	doi = {10.1109/ICRA.2016.7487492},
	language = {en},
	urldate = {2021-03-01},
	booktitle = {2016 {IEEE} {International} {Conference} on {Robotics} and {Automation} ({ICRA})},
	publisher = {IEEE},
	author = {Rosen, Michelle H. and le Pivain, Geoffroy and Sahai, Ranjana and Jafferis, Noah T. and Wood, Robert J.},
	month = may,
	year = {2016},
	pages = {3227--3233}
}

@article{wissa_free_2015,
	title = {Free {Flight} {Testing} and {Performance} {Evaluation} of a {Passively} {Morphing} {Ornithopter}},
	volume = {7},
	issn = {1756-8293, 1756-8307},
	url = {http://journals.sagepub.com/doi/10.1260/1756-8293.7.1.21},
	doi = {10.1260/1756-8293.7.1.21},
	language = {en},
	number = {1},
	urldate = {2021-03-01},
	journal = {International Journal of Micro Air Vehicles},
	author = {Wissa, Aimy and Grauer, Jared and Guerreiro, Nelson and Hubbard, James and Altenbuchner, Cornelia and Tummala, Yashwanth and Frecker, Mary and Roberts, Richard},
	month = mar,
	year = {2015},
	pages = {21--40}
}

@article{ramezani_biomimetic_2017,
	title = {A biomimetic robotic platform to study flight specializations of bats},
	volume = {2},
	issn = {2470-9476},
	url = {https://robotics.sciencemag.org/lookup/doi/10.1126/scirobotics.aal2505},
	doi = {10.1126/scirobotics.aal2505},
	language = {en},
	number = {3},
	urldate = {2021-03-01},
	journal = {Sci. Robot.},
	author = {Ramezani, Alireza and Chung, Soon-Jo and Hutchinson, Seth},
	month = feb,
	year = {2017},
	pages = {eaal2505}
}

@inproceedings{ramezani_bat_2016,
	address = {Stockholm, Sweden},
	title = {Bat {Bot} ({B2}), a biologically inspired flying machine},
	isbn = {978-1-4673-8026-3},
	url = {http://ieeexplore.ieee.org/document/7487491/},
	doi = {10.1109/ICRA.2016.7487491},
	language = {en},
	urldate = {2021-03-01},
	booktitle = {2016 {IEEE} {International} {Conference} on {Robotics} and {Automation} ({ICRA})},
	publisher = {IEEE},
	author = {Ramezani, Alireza and Shi, Xichen and Chung, Soon-Jo and Hutchinson, Seth},
	month = may,
	year = {2016},
	pages = {3219--3226}
}

@incollection{mangan_describing_2017,
	address = {Cham},
	title = {Describing {Robotic} {Bat} {Flight} with {Stable} {Periodic} {Orbits}},
	volume = {10384},
	isbn = {978-3-319-63536-1 978-3-319-63537-8},
	url = {http://link.springer.com/10.1007/978-3-319-63537-8_33},
	language = {en},
	urldate = {2021-03-01},
	booktitle = {Biomimetic and {Biohybrid} {Systems}},
	publisher = {Springer International Publishing},
	author = {Ramezani, Alireza and Ahmed, Syed Usman and Hoff, Jonathan and Chung, Soon-Jo and Hutchinson, Seth},
	editor = {Mangan, Michael and Cutkosky, Mark and Mura, Anna and Verschure, Paul F.M.J. and Prescott, Tony and Lepora, Nathan},
	year = {2017},
	doi = {10.1007/978-3-319-63537-8_33},
	note = {Series Title: Lecture Notes in Computer Science},
	pages = {394--405}
}

@inproceedings{hoff_synergistic_2016,
	title = {Synergistic {Design} of a {Bio}-{Inspired} {Micro} {Aerial} {Vehicle} with {Articulated} {Wings}},
	isbn = {978-0-9923747-2-3},
	url = {http://www.roboticsproceedings.org/rss12/p09.pdf},
	doi = {10.15607/RSS.2016.XII.009},
	language = {en},
	urldate = {2021-03-01},
	booktitle = {Robotics: {Science} and {Systems} {XII}},
	publisher = {Robotics: Science and Systems Foundation},
	author = {Hoff, Jonathan and Ramezani, Alireza and Chung, Soon-Jo and Hutchinson, Seth},
	year = {2016}
}

@inproceedings{ramezani_lagrangian_2015,
	address = {Hamburg, Germany},
	title = {Lagrangian modeling and flight control of articulated-winged bat robot},
	isbn = {978-1-4799-9994-1},
	url = {http://ieeexplore.ieee.org/document/7353772/},
	doi = {10.1109/IROS.2015.7353772},
	language = {en},
	urldate = {2021-03-01},
	booktitle = {2015 {IEEE}/{RSJ} {International} {Conference} on {Intelligent} {Robots} and {Systems} ({IROS})},
	publisher = {IEEE},
	author = {Ramezani, Alireza and Shi, Xichen and Chung, Soon-Jo and Hutchinson, Seth},
	month = sep,
	year = {2015},
	pages = {2867--2874}
}

@article{hoff_optimizing_2018,
	title = {Optimizing the structure and movement of a robotic bat with biological kinematic synergies},
	volume = {37},
	issn = {0278-3649, 1741-3176},
	url = {http://journals.sagepub.com/doi/10.1177/0278364918804654},
	doi = {10.1177/0278364918804654},
	language = {en},
	number = {10},
	urldate = {2021-03-01},
	journal = {The International Journal of Robotics Research},
	author = {Hoff, Jonathan and Ramezani, Alireza and Chung, Soon-Jo and Hutchinson, Seth},
	month = sep,
	year = {2018},
	pages = {1233--1252}
}

@incollection{mangan_reducing_2017,
	address = {Cham},
	title = {Reducing {Versatile} {Bat} {Wing} {Conformations} to a 1-{DoF} {Machine}},
	volume = {10384},
	isbn = {978-3-319-63536-1 978-3-319-63537-8},
	url = {http://link.springer.com/10.1007/978-3-319-63537-8_16},
	language = {en},
	urldate = {2021-03-01},
	booktitle = {Biomimetic and {Biohybrid} {Systems}},
	publisher = {Springer International Publishing},
	author = {Hoff, Jonathan and Ramezani, Alireza and Chung, Soon-Jo and Hutchinson, Seth},
	editor = {Mangan, Michael and Cutkosky, Mark and Mura, Anna and Verschure, Paul F.M.J. and Prescott, Tony and Lepora, Nathan},
	year = {2017},
	doi = {10.1007/978-3-319-63537-8_16},
	note = {Series Title: Lecture Notes in Computer Science},
	pages = {181--192}
}

@inproceedings{hoff_trajectory_2019,
	address = {Macau, China},
	title = {Trajectory planning for a bat-like flapping wing robot},
	isbn = {978-1-72814-004-9},
	url = {https://ieeexplore.ieee.org/document/8968450/},
	doi = {10.1109/IROS40897.2019.8968450},
	language = {en},
	urldate = {2021-03-01},
	booktitle = {2019 {IEEE}/{RSJ} {International} {Conference} on {Intelligent} {Robots} and {Systems} ({IROS})},
	publisher = {IEEE},
	author = {Hoff, Jonathan and Syed, Usman and Ramezani, Alireza and Hutchinson, Seth},
	month = nov,
	year = {2019},
	pages = {6800--6805}
}

@article{sihite_computational_2020,
	title = {Computational {Structure} {Design} of a {Bio}-{Inspired} {Armwing} {Mechanism}},
	volume = {5},
	issn = {2377-3766, 2377-3774},
	url = {https://ieeexplore.ieee.org/document/9143405/},
	doi = {10.1109/LRA.2020.3010217},
	language = {en},
	number = {4},
	urldate = {2021-03-01},
	journal = {IEEE Robot. Autom. Lett.},
	author = {Sihite, Eric and Kelly, Peter and Ramezani, Alireza},
	month = oct,
	year = {2020},
	pages = {5929--5936}
}

@article{sihite_integrated_2021,
	title = {An {Integrated} {Mechanical} {Intelligence} and {Control} {Approach} {Towards} {Flight} {Control} of {Aerobat}},
	url = {https://arxiv.org/abs/2103.16566v1},
	language = {en},
	urldate = {2021-04-11},
	author = {Sihite, Eric and Darabi, Atefe and Dangol, Pravin and Lessieur, Andrew and Ramezani, Alireza},
	month = mar,
	year = {2021},
}

@inproceedings{sihite_enforcing_2020,
	address = {Jeju Island, Korea (South)},
	title = {Enforcing nonholonomic constraints in {Aerobat}, a roosting flapping wing model},
	isbn = {978-1-72817-447-1},
	url = {https://ieeexplore.ieee.org/document/9304158/},
	doi = {10.1109/CDC42340.2020.9304158},
	language = {en},
	urldate = {2021-03-01},
	booktitle = {2020 59th {IEEE} {Conference} on {Decision} and {Control} ({CDC})},
	publisher = {IEEE},
	author = {Sihite, Eric and Ramezani, Alireza},
	month = dec,
	year = {2020},
	pages = {5321--5327}
}

@article{boutet_unsteady_2018,
	title = {Unsteady {Lifting} {Line} {Theory} {Using} the {Wagner} {Function} for the {Aerodynamic} and {Aeroelastic} {Modeling} of {3D} {Wings}},
	volume = {5},
	issn = {2226-4310},
	url = {http://www.mdpi.com/2226-4310/5/3/92},
	doi = {10.3390/aerospace5030092},
	language = {en},
	number = {3},
	urldate = {2021-03-01},
	journal = {Aerospace},
	author = {Boutet, Johan and Dimitriadis, Grigorios},
	month = sep,
	year = {2018},
	pages = {92}
}

@article{izraelevitz_state-space_2017,
	title = {State-{Space} {Adaptation} of {Unsteady} {Lifting} {Line} {Theory}: {Twisting}/{Flapping} {Wings} of {Finite} {Span}},
	volume = {55},
	issn = {0001-1452, 1533-385X},
	shorttitle = {State-{Space} {Adaptation} of {Unsteady} {Lifting} {Line} {Theory}},
	url = {https://arc.aiaa.org/doi/10.2514/1.J055144},
	doi = {10.2514/1.J055144},
	language = {en},
	number = {4},
	urldate = {2021-03-01},
	journal = {AIAA Journal},
	author = {Izraelevitz, Jacob S. and Zhu, Qiang and Triantafyllou, Michael S.},
	month = apr,
	year = {2017},
	pages = {1279--1294}
}

@article{phan_insect-inspired_2019,
	title = {Insect-inspired, tailless, hover-capable flapping-wing robots: {Recent} progress, challenges, and future directions},
	volume = {111},
	issn = {03760421},
	shorttitle = {Insect-inspired, tailless, hover-capable flapping-wing robots},
	url = {https://linkinghub.elsevier.com/retrieve/pii/S0376042119300545},
	doi = {10.1016/j.paerosci.2019.100573},
	language = {en},
	urldate = {2021-03-01},
	journal = {Progress in Aerospace Sciences},
	author = {Phan, Hoang Vu and Park, Hoon Cheol},
	month = nov,
	year = {2019},
	pages = {100573}
}

@article{peterson_wing-assisted_2011,
	title = {A wing-assisted running robot and implications for avian flight evolution},
	volume = {6},
	issn = {1748-3182, 1748-3190},
	url = {https://iopscience.iop.org/article/10.1088/1748-3182/6/4/046008},
	doi = {10.1088/1748-3182/6/4/046008},
	language = {en},
	number = {4},
	urldate = {2021-03-01},
	journal = {Bioinspir. Biomim.},
	author = {Peterson, K and Birkmeyer, P and Dudley, R and Fearing, R S},
	month = dec,
	year = {2011},
	pages = {046008}
}

@article{chukewad_robofly_2020,
	title = {{RoboFly}: {An} insect-sized robot with simplified fabrication that is capable of flight, ground, and water surface locomotion},
	shorttitle = {{RoboFly}},
	url = {http://arxiv.org/abs/2001.02320},
	language = {en},
	urldate = {2021-03-01},
	journal = {arXiv:2001.02320 [cs, eess]},
	author = {Chukewad, Yogesh M. and James, Johannes and Singh, Avinash and Fuller, Sawyer},
	month = oct,
	year = {2020},
	note = {arXiv: 2001.02320}
}

@article{ma_controlled_2013,
	title = {Controlled {Flight} of a {Biologically} {Inspired}, {Insect}-{Scale} {Robot}},
	volume = {340},
	issn = {0036-8075, 1095-9203},
	url = {https://www.sciencemag.org/lookup/doi/10.1126/science.1231806},
	doi = {10.1126/science.1231806},
	language = {en},
	number = {6132},
	urldate = {2021-03-01},
	journal = {Science},
	author = {Ma, K. Y. and Chirarattananon, P. and Fuller, S. B. and Wood, R. J.},
	month = may,
	year = {2013},
	pages = {603--607}
}

@article{hauser_role_2012,
	title = {The role of feedback in morphological computation with compliant bodies},
	volume = {106},
	issn = {0340-1200, 1432-0770},
	url = {http://link.springer.com/10.1007/s00422-012-0516-4},
	doi = {10.1007/s00422-012-0516-4},
	language = {en},
	number = {10},
	urldate = {2021-03-01},
	journal = {Biol Cybern},
	author = {Hauser, Helmut and Ijspeert, Auke J. and Füchslin, Rudolf M. and Pfeifer, Rolf and Maass, Wolfgang},
	month = nov,
	year = {2012},
	pages = {595--613}
}

@article{send_artificial_2012,
	title = {Artificial hinged-wing bird with active torsion and partially linear kinematics},
	language = {en},
	journal = {Proceeding of 28th Congress of the International Council of the Aeronautical Sciences},
	author = {Send, Wolfgang and Fischer, Markus and Jebens, Kristof and Mugrauer, Rainer and Nagarathinam, Agalya and Scharstein, Felix},
	year = {2012},
	pages = {10}
}

@article{farrell_helbling_review_2018,
	title = {A {Review} of {Propulsion}, {Power}, and {Control} {Architectures} for {Insect}-{Scale} {Flapping}-{Wing} {Vehicles}},
	volume = {70},
	issn = {0003-6900, 2379-0407},
	url = {https://asmedigitalcollection.asme.org/appliedmechanicsreviews/article/doi/10.1115/1.4038795/443695/A-Review-of-Propulsion-Power-and-Control},
	doi = {10.1115/1.4038795},
	language = {en},
	number = {1},
	urldate = {2021-03-02},
	journal = {Applied Mechanics Reviews},
	author = {Farrell Helbling, E. and Wood, Robert J.},
	month = jan,
	year = {2018},
	pages = {010801}
}

@article{tu_untethered_2020,
	title = {Untethered {Flight} of an {At}-{Scale} {Dual}-motor {Hummingbird} {Robot} with {Bio}-inspired {Decoupled} {Wings}},
	issn = {2377-3766, 2377-3774},
	url = {https://ieeexplore.ieee.org/document/9001181/},
	doi = {10.1109/LRA.2020.2974717},
	language = {en},
	urldate = {2021-03-02},
	journal = {IEEE Robot. Autom. Lett.},
	author = {Tu, Zhan and Fei, Fan and Deng, Xinyan},
	year = {2020},
	pages = {1--1}
}

@article{hubel_wake_2010,
	title = {Wake structure and wing kinematics: the flight of the lesser dog-faced fruit bat, Cynopterus brachyotis},
	volume = {213},
	issn = {0022-0949},
	url = {https://doi.org/10.1242/jeb.043257},
	doi = {10.1242/jeb.043257},
	shorttitle = {Wake structure and wing kinematics},
	pages = {3427--3440},
	number = {20},
	journaltitle = {Journal of Experimental Biology},
	author = {Hubel, Tatjana Y. and Riskin, Daniel K. and Swartz, Sharon M. and Breuer, Kenneth S.},
	urldate = {2021-08-04},
	date = {2010-10-15}
}

@inproceedings{willis2007computational,
  title={A computational framework for fluid structure interaction in biologically inspired flapping flight},
  author={Willis, David and Israeli, Emily and Persson, Per-Olof and Drela, Mark and Peraire, Jaime and Swartz, Sharon and Breuer, Kenny},
  booktitle={25th AIAA Applied Aerodynamics Conference},
  pages={3803},
  year={2007}
}

@article{riskin_quantifying_2008,
	title = {Quantifying the complexity of bat wing kinematics},
	volume = {254},
	issn = {00225193},
	url = {https://linkinghub.elsevier.com/retrieve/pii/S002251930800324X},
	doi = {10.1016/j.jtbi.2008.06.011},
	language = {en},
	number = {3},
	urldate = {2021-03-02},
	journal = {Journal of Theoretical Biology},
	author = {Riskin, Daniel K. and Willis, David J. and Iriarte-Díaz, José and Hedrick, Tyson L. and Kostandov, Mykhaylo and Chen, Jian and Laidlaw, David H. and Breuer, Kenneth S. and Swartz, Sharon M.},
	month = oct,
	year = {2008},
	pages = {604--615}
}

@article{gerdes_robo_2014,
	title = {Robo {Raven}: {A} {Flapping}-{Wing} {Air} {Vehicle} with {Highly} {Compliant} and {Independently} {Controlled} {Wings}},
	volume = {1},
	issn = {2169-5172, 2169-5180},
	shorttitle = {Robo {Raven}},
	url = {https://www.liebertpub.com/doi/10.1089/soro.2014.0019},
	doi = {10.1089/soro.2014.0019},
	language = {en},
	number = {4},
	urldate = {2021-04-11},
	journal = {Soft Robotics},
	author = {Gerdes, John and Holness, Alex and Perez-Rosado, Ariel and Roberts, Luke and Greisinger, Adrian and Barnett, Eli and Kempny, Johannes and Lingam, Deepak and Yeh, Chen-Haur and Bruck, Hugh A. and Gupta, Satyandra K.},
	month = dec,
	year = {2014},
	pages = {275--288}
}

@article{parslew2013theoretical,
  title={Theoretical modelling of wakes from retractable flapping wings in forward flight},
  author={Parslew, Ben and Crowther, William J},
  journal={PeerJ},
  volume={1},
  pages={e105},
  year={2013},
  publisher={PeerJ Inc.}
}

@article{hedenstrom_bat_2015,
	title = {Bat flight: aerodynamics, kinematics and flight morphology},
	volume = {218},
	issn = {0022-0949, 1477-9145},
	url = {http://jeb.biologists.org/cgi/doi/10.1242/jeb.031203},
	doi = {10.1242/jeb.031203},
	shorttitle = {Bat flight},
	pages = {653--663},
	number = {5},
	journaltitle = {Journal of Experimental Biology},
	author = {Hedenstrom, A. and Johansson, L. C.},
	urldate = {2020-11-24},
	date = {2015-03-01},
	langid = {english}
}

\end{document}